\documentclass[conference]{IEEEtran}
\IEEEoverridecommandlockouts
\usepackage{cite}
\usepackage{amsmath,amssymb,amsfonts}
\usepackage{algorithmic}
\usepackage{graphicx}
\usepackage{textcomp}
\usepackage{xcolor}
\def\BibTeX{{\rm B\kern-.05em{\sc i\kern-.025em b}\kern-.08em
    T\kern-.1667em\lower.7ex\hbox{E}\kern-.125emX}}

\usepackage{listings}
\usepackage{xcolor}
\usepackage{caption}

\usepackage{siunitx}
\usepackage{tikz}
\usetikzlibrary{shapes, positioning}

\usepackage{tabularx}
\usepackage{booktabs}

\usepackage{url}
    
\begin{document}

\title{Feedback-Induced Performance Decline in LLM-Based Decision-Making}

\author{
\IEEEauthorblockN{Xiao Yang}
\IEEEauthorblockA{\textit{Dept. of ECSE}, \textit{Monash University}\\
Clayton, Australia}
\and
\IEEEauthorblockN{Juxi Leitner}
\IEEEauthorblockA{\textit{Dept. of ECSE}, \textit{Monash University}\\
Clayton, Australia}
\and
\IEEEauthorblockN{Michael Burke}
\IEEEauthorblockA{\textit{Dept. of ECSE}, \textit{Monash University}\\
Clayton, Australia}
}

\maketitle

\begin{abstract}
The ability of Large Language Models (LLMs) to extract context from natural language problem descriptions naturally raises questions about their suitability in autonomous decision-making settings. This paper studies the behaviour of these models  within a Markov Decision Process (MDPs). While traditional reinforcement learning (RL) strategies commonly employed in this setting rely on iterative exploration, LLMs, pre-trained on diverse datasets, offer the capability to leverage prior knowledge for faster adaptation. We investigate online structured prompting strategies in sequential decision making tasks, comparing the zero-shot performance of LLM-based approaches to that of classical RL methods. Our findings reveal that although LLMs demonstrate improved initial performance in simpler environments, they struggle with planning and reasoning in complex scenarios without fine-tuning or additional guidance. Our results show that feedback mechanisms, intended to improve decision-making, often introduce confusion, leading to diminished performance in intricate environments. These insights underscore the need for further exploration into hybrid strategies, fine-tuning, and advanced memory integration to enhance LLM-based decision-making capabilities.
\end{abstract}


\section{Introduction}

The integration of Large Language Models (LLMs) into complex decision-making tasks is promising, given their ability to process and generate text in a manner that hints at human-like reasoning. This research explores the potential of LLMs as decision-making agents in Markov Decision Processes (MDPs)—a formal framework for sequential decision-making under uncertainty. While traditional reinforcement learning (RL) methods have proven effective at learning optimal policies within this framework, they rely heavily on sample-inefficient iterative exploration and often struggle in domains where real-time adaptability and contextual understanding are critical. In contrast, LLMs pretrained on diverse datasets may offer a unique advantage by leveraging prior knowledge to support faster adaptation and more effective reasoning.

In this study, we investigate the use of structured prompting strategies to guide LLMs in solving MDPs, focusing on how varying levels of contextual conditioning—ranging from state-only prompts to those incorporating memory and reward feedback—affect decision-making performance. By leveraging the flexibility of prompts and the reasoning capabilities of LLMs, we initially hypothesized that LLM-based policies could outperform traditional RL algorithms in specific scenarios. However, the results revealed otherwise: on their own, LLMs cannot effectively plan or reason, and performance declines with additional context when naive prompting strategies are applied.

The primary contributions of this work are as follows:
\begin{itemize}
\item A comprehensive evaluation of LLM-based policies across multiple configurations of the MiniGrid environment, assessing their performance relative to classical RL methods.
\item Demonstrating that while LLMs are equipped with extensive prior knowledge, they lack sufficient grounding and reasoning skills to effectively utilize this knowledge for problem-solving without additional guidance.
\item Revealing that the incorporation of feedback can lead to policy degradation, where extraneous or misaligned information detracts from the model’s decision-making effectiveness.
\end{itemize}

\section{Related Work}

Despite advances in prompting and hybrid methods, LLMs continue to struggle to produce reliable and executable plans in complex domains. Kambhampati et al. \cite{darwiche2024llms,valmeekam2023planning,valmeekam2023planbench,guan2023leveraging} have argued that LLMs are fundamentally incapable of true planning or reasoning. They contend that while fine-tuning may improve empirical reasoning performance, it effectively transforms planning tasks into memory-based approximate retrieval problems rather than fostering genuine planning capabilities. Furthermore, Kambhampati et al. \cite{kambhampati2024position} demonstrate that although LLMs can generate plausible planning sequences, closer inspection often reveals these sequences to be invalid. They emphasize that LLMs excel at information extraction but lack true planning competence. Consequently, external verifiers or symbolic planners are often required to ensure the feasibility and correctness of LLM-generated outputs. Huang et al. \cite{huang2022language} similarly show that although LLM-generated planning sequences are often rated highly by human evaluators, they frequently fail in actual execution. To address this gap, they propose using a separate model to translate natural language plans from LLMs into a constrained vocabulary. While this approach significantly improves executability, it also reduces human evaluation scores—likely due to the limited expressiveness of the constrained vocabulary. The emergence of reasoning-focused models—often referred to as Large Reasoning Models (LRMs)—trained via reinforcement learning to enhance chain-of-thought (CoT) reasoning \cite{raschka2025understanding,guo2025deepseek}, has led to some improvements. However, recent studies \cite{kambhampati2025stop,stechly2024self,shojaee2025illusion} indicate that while LRMs perform well in domains such as mathematics, where well-defined external verifiers exist, their reasoning capabilities do not generalize well to other domains. These limitations are often rooted in training dependencies and further support the view that both LLMs and LRMs rely primarily on memory retrieval rather than genuine reasoning. While these findings align with our own observations, this study specifically examines the planning capabilities of LLMs within the context of Markov Decision Processes (MDPs). Relatively little attention has been given to how feedback mechanisms—such as state transitions, reward signals, or policy traces—influence the ability of LLMs to solve sequential MDPs. This work investigates whether LLMs are capable of solving MDP problems independently.

A central strategy for improving LLM performance in decision-making tasks is prompt engineering. Techniques such as Chain-of-Thought (CoT) prompting \cite{wei2022chain} and ReAct \cite{yao2023react} demonstrate how structured prompts can guide models toward more coherent and interpretable reasoning by incorporating intermediate steps. However, these methods have primarily been applied to static or single-step problems rather than to sequential settings characterized by dynamic feedback loops. More recent developments, including DSPy \cite{khattab2023dspy} and TextGrad \cite{yuksekgonul2024textgrad}, aim to automate prompt refinement, thereby alleviating the burden of manual prompt construction. While these methods are not directly employed in this study, they underscore the growing significance of prompt optimization for real-world applications.

Reinforcement learning (RL) has been used extensively to solve MDP problems. One popular RL algorithm is Proximal Policy Optimization (PPO), which updates policies in a stable and efficient manner by constraining policy changes through a clipped objective function, thereby balancing exploration and exploitation \cite{schulman2017proximal}. Although RL can handle highly complex MDP problems, it usually requires a significant amount of exploration. Recent studies have explored combining LLMs with RL. In these approaches, LLMs typically serve as high-level policy planners or skill selectors, while RL components adaptively refine behaviors based on environmental feedback \cite{prakash2023llm,tang2024mba}. These methods have shown that LLMs can introduce valuable priors in exploration-heavy domains. However, they often require model updates or fine-tuning, which contrasts with the zero-shot, prompt-based paradigm investigated in this study.

To address these challenges, several works have proposed combining LLMs with external search or solver tools. For example, \cite{schultz2024mastering} integrates LLMs with Monte Carlo Tree Search (MCTS) to achieve strategic proficiency in board games, highlighting the synergy between neural reasoning and systematic search. Similarly, \cite{ju2024globe} demonstrates how pairing LLMs with linear programming solvers can yield optimal solutions in travel planning problems. Huang \cite{huang2025ai} further shows that AI systems, exemplified by GitHub Copilot, can autonomously generate formal TLA+ specifications from large-scale production codebases, uncovering subtle concurrency bugs—such as race conditions in Azure Storage—using a TLA+ verifier. In contrast, our study deliberately avoids such integrations, focusing instead on evaluating the intrinsic capabilities of LLMs as decision-making agents within MDPs, without assistance from external verifiers or solvers. While we employ a simulation environment with feedback-based prompting—which may be viewed as a form of weak verification—the primary focus is to assess whether LLMs can reason effectively within MDP settings.

\section{Methodology}

\subsection{Problem Formulation}

We consider sequential decision making tasks that can be described as a Markov decision making process. At each step, an agent receives an observation of the environment and agent state $s_t \in S$ and selects an action $a_t \in A$ that affects the environment's state. The environment provides feedback in the form of state transitions following some transition probability $P(s_t|s_{t-1},a_t)$ and rewards $r_t$ generated using some reward function $R(s_t,a_t)$. The agent's objective is to find a policy $\pi$ or mapping from states to actions that maximizes cumulative rewards over time. We consider settings where the transition dynamics are unknown, and we do not have knowledge of reward function.

\subsection{LLM Policies}\label{sec:llm_policies}

For our experiments, the policy $\pi$ is generated by a Large Language Model (LLM) interacting with the environment. The LLM-based policy is defined by various prompting strategies, categorised as follows:

\begin{itemize}
    \item \emph{State-Only Policy} ($\pi_{\text{LLM}}(s_t)$): The LLM is conditioned solely on the current state $s_t$, by encoding this into a suitable text description. This model relies on implicit knowledge to infer optimal actions without additional context about previous interactions.
    
    \item \emph{State and Memory Policy} ($\pi_{\text{LLM}}(s_t, m_t)$): The policy is conditioned on an encoding of the current state $s_t$ and a memory component $m_t$ into text, where the memory component  includes historical states and actions, or summaries of prior policy information, potentially enabling more informed decision-making.
    
    \item \emph{State, Memory, and Reward Policy} ($\pi_{\text{LLM}}(s_t, m_t, r_t$): This configuration incorporates immediate reward feedback $r_t$ allowing the LLM to adapt based on recent outcomes.
\end{itemize}

Our hypothesis is that LLMs, pre-trained on diverse datasets, can leverage prior knowledge through prompting to navigate MDPs effectively, in contrast to traditional reinforcement learning (RL) algorithms, which require lengthy iterative exploration during training. To infer $\pi$, we employ structured prompts that provide varying levels of context, including information about states, memory, and rewards. By evaluating these prompts, we aim to understand how contextual conditioning impacts the LLM's decision-making capabilities.  


\begin{table*}[t]
\centering
\begin{tabular}{lccc}
\toprule
Approach & State & Memory & Reward \\
& $s_t$ & $m_t$ & $r_t$ \\
\midrule
HWBP & * & & \\
HWBP + CoT & * & &\\
HWBP + CoT + DF & * & * &\\
HWBP + CoT + DF + RF & * & * & *\\
HWBP + CoT + DF + RF + CRF & * & * & *\\
HWBP + CoT + DF + RF + CRF + PF & * & * & *\\
HWBP + CoT + DF + RF + CRF + PF across episodes & * & * & *\\
\bottomrule
\end{tabular}
\caption{Comparison of LLM prompting strategies based on whether they incorporate state ($s_t$), memory ($m_t$), and/or reward ($r_t$) information.}
\label{tab:llm-policy-types}
\end{table*}

We evaluate several prompt-based strategies to enhance LLM decision-making within Markov Decision Process (MDP) settings, comparing them against two baselines: a random policy and a PPO-trained reinforcement learning (RL) policy. The simplest LLM approach uses a human-written base prompt (HWBP) describing the task. This is incrementally extended by incorporating chain-of-thought (CoT) reasoning, dynamics feedback (DF) from action-state transitions, reward feedback (RF), cumulative reward feedback (CRF), and policy feedback (PF), where the LLM summarizes and reuses its own strategy. The most advanced configuration carries policy feedback across episodes to potentially support learning over time. These approaches collectively aim to provide the LLM with richer contextual and feedback signals, as summarized in Table~\ref{tab:llm-policy-types}, enabling progressively more informed and adaptive behavior.

\subsection{Experimental Settings}\label{sec:exp_settings}

\begin{figure*}[t]
    \centering
    \includegraphics[width=0.28\textwidth]{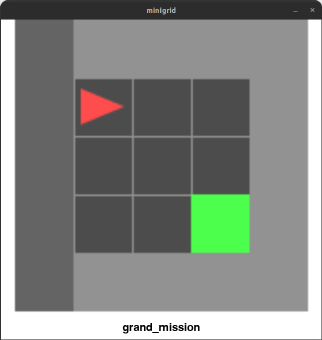}
    \includegraphics[width=0.28\textwidth]{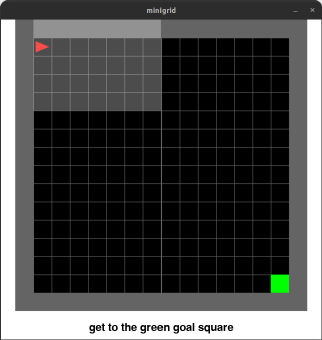}
    \includegraphics[width=0.28\textwidth]{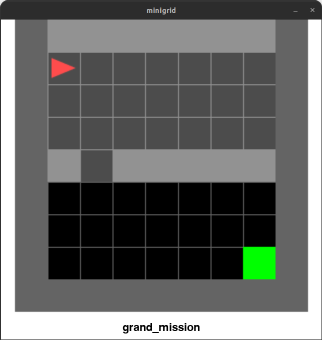}
    \caption{MiniGrid configuration environments used for testing (left 5x5 grid, middle 16x16 grid, right 9x9 board with an internal partition wall. The llm policy needs to sequentially generate actions to move the agent (red arrow) to the green square.}
    \label{fig:minigrid-configs}
\end{figure*}

Experiments were conducted within the MiniGrid environment \cite{chevalier2023minigrid} using three configurations of varying complexity. The MiniGrid environment provides a grid-based world where an agent must navigate to a goal while avoiding obstacles. Figure~\ref{fig:minigrid-configs} illustrates the initial states of the different MiniGrid configurations used in our experiments. The agent, depicted as a red triangle, aims to reach the target represented by a green square, with walls serving as the primary obstacles. For approaches utilizing Large Language Models (LLMs), the environment is encoded in a text format to facilitate prompt generation; additional details on this text representation are provided in Appendix~\ref{app:src}.
    
We employ Llama~3.1~8B and Qwen~2.5~1.5b accessed via the Ollama API~\cite{ollama_llama3_2024,grattafiori2024llama,ollama_qwen25_2024,yang2025qwen3} as the underlying language models for our experiments, using the following parameters and evaluation metrics.

Each approach outlined  above is evaluated over 100 episodes. Episodes are terminated if the agent fails to complete the task within a predefined step limit. We set this limit to 100 steps, which is sufficiently larger than the optimal path length, providing ample opportunity for the agent to reach the goal.

We evaluate policies using Cumulative Reward per Episode, which tracks the total reward accumulated by the agent in each episode. We also report the success rate over episodes, the percentage of episodes in which the agent reaches the goal within the step limit.

\subsection{One-shot Prompting with Reasoning Models}\label{sec:rm}

To evaluate the effectiveness of sequential decision-making with LLM-based policies, we also leverage recent advancements in reasoning models. Two local models—Deepseek R1 14B \cite{ollama_deepseekr114b_2025,guo2025deepseek} and QwQ 32B \cite{ollama_qwq32b_2025,yang2025qwen3}—were used for comparison with the earlier LLM-based policies. Since reasoning models are trained to produce step-by-step solutions, we applied one-shot prompting to generate complete planned solutions from the initial states for all three configurations. The outputs were then evaluated for quality.

\section{Results}

\begin{figure*}[ht]
\centering
\includegraphics[width=0.85\textwidth]{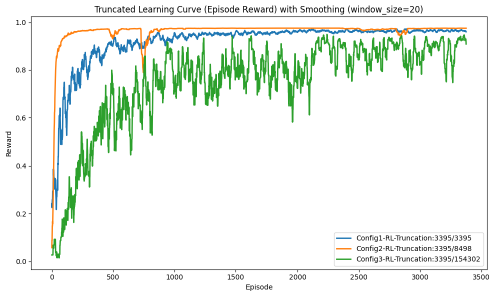}
\caption{RL learning curves for Configurations 1, 2, and 3. The curves are truncated to the same number of episodes for illustration purposes. Note that, due to differences in problem size, the number of episodes does not directly correspond to the number of training steps (see appendix \ref{app:rl_hyper}).
}
\label{fig:learning_curve_rl}
\end{figure*}

A random policy and reinforcement learning (RL) are used as baseline approaches. As with the LLM-based policies, standard evaluation metrics are applied. For RL, learning curves are shown in Figure~\ref{fig:learning_curve_rl} to illustrate training progress across different configurations. Notably, all LLM-based policies—except one—as well as the random policy, do not involve any training or learning phases. The only LLM-based policy that includes a learning phase is the one that retains policy feedback across episodes. However, since its rewards remain approximately zero across all three configurations, it demonstrates no effective learning (see Tables~\ref{tab:merged-config1-results}, \ref{tab:merged-config2-results}, and \ref{tab:merged-config3-results}); therefore, its learning curves are omitted from Figure~\ref{fig:learning_curve_rl}. Table~\ref{tab:rl_shared_params} and Table \ref{tab:rl_differences} in appendix \ref{app:rl_hyper} summarize the hyperparameters used to train the RL policies. 

As shown in the results tables, the RL policies achieved 100\% success rates and average rewards above 0.9 across all configurations, while the random policy performed significantly worse, with average rewards of 0.216, 0.047, and 0.015 for Configurations 1, 2, and 3, respectively. Interestingly, LLM-based policies that incorporate policy feedback can perform worse than the random policy when the underlying LLM is LLaMA 3.1 8B.

\subsection{Configuration 1: 5x5 Grid without Internal Obstacles}

This configuration represents the simplest environment with minimal complexity. The performance metrics for each approach are summarized in Table~\ref{tab:merged-config1-results}. In general, while a good human written base prompt performs substantially better than random, we see increasing performance decline as additional forms of feedback are provided to both LLMs. None of the LLM-based approaches can match the performance of the baseline RL. Interestingly, Qwen2.5:1.5B—despite having less than 20\% of the parameters of LLaMA3.1:8B—can outperform the latter, although the overall trend remains similar.

\begin{table*}[t]
    \centering
    \caption{Performance Metrics for Configuration~1.}
    \label{tab:merged-config1-results}
    \begin{tabular}{lSSSS}
        \toprule
        \textbf{Reference Approach} & \textbf{Avg. Cum. Reward} & \textbf{Success Rate (\%)} & & \\ 
        \midrule
        Random Policy & 0.216 & 44 & & \\
        RL Baseline & 0.950 & 100 & & \\ 
        \midrule
        \textbf{LLM-based Approach} &
        \multicolumn{2}{c}{\textbf{llama3.1:8b}} &
        \multicolumn{2}{c}{\textbf{qwen2.5:1.5b}} \\
        \cmidrule(lr){2-3} \cmidrule(lr){4-5}
        & {\textbf{Avg. Cum. Reward}} & {\textbf{Success Rate (\%)}} 
        & {\textbf{Avg. Cum. Reward}} & {\textbf{Success Rate (\%)}} \\
        \midrule
        HWBP & 0.748 & 98 & 0.589 & 80 \\
        HWBP + CoT & 0.743 & 98 & 0.681 & 92 \\
        HWBP + CoT + DF & 0.714 & 96 & 0.000 & 0 \\
        HWBP + CoT + DF + RF & 0.637 & 93 & 0.41 & 63 \\
        HWBP + CoT + DF + RF + CRF & 0.677 & 95 & 0.392 & 61 \\
        HWBP + CoT + DF + RF + CRF + PF & 0.170 & 33 & 0.568 & 88 \\
        Prev config + PF across episodes & 0.030 & 9 & {--} & {--} \\
        HWBP + CRF & 0.69 & 96 & 0.57 & 86 \\
        HWBP + PF & 0 & 0 & 0.559 & 81 \\
        \bottomrule
    \end{tabular}
\end{table*}

\subsection{Configuration 2: 16x16 Grid without Internal Obstacles}

This configuration introduced a larger grid size, increasing the environment's complexity. Results are presented in Table \ref{tab:merged-config2-results}. As before, there is noticeable performance decline as additional forms of feedback are provided, though a good human written base prompt performed relatively well with a qwen2.5:1.5b model.

\begin{table*}[t]
    \centering
    \caption{Performance Metrics for Configuration~2.}
    \label{tab:merged-config2-results}
    \begin{tabular}{lSSSS}
        \toprule
        \textbf{Reference Approach} & \textbf{Avg. Cum. Reward} & \textbf{Success Rate (\%)} & & \\ 
        \midrule
        Random Policy & 0.047 & 11 &  &  \\
        RL Baseline & 0.920 & 100 &  &  \\ 
        \midrule
        \textbf{LLM-based Approach} &
        \multicolumn{2}{c}{\textbf{llama3.1:8b}} &
        \multicolumn{2}{c}{\textbf{qwen2.5:1.5b}} \\
        \cmidrule(lr){2-3} \cmidrule(lr){4-5}
        & {\textbf{Avg. Cum. Reward}} & {\textbf{Success Rate (\%)}} 
        & {\textbf{Avg. Cum. Reward}} & {\textbf{Success Rate (\%)}} \\
        \midrule
        HWBP & 0.728 & 98 & 0.815 & 100 \\
        HWBP + CoT & 0.660 & 94 & 0.735 & 97 \\
        HWBP + CoT + DF & 0.315 & 64 & 0.734 & 82 \\
        HWBP + CoT + DF + RF & 0.313 & 63 & 0.742 & 89 \\
        HWBP + CoT + DF + RF + CRF & 0.448 & 79 & 0.776 & 99 \\
        HWBP + CoT + DF + RF + CRF + PF & 0.000 & 0 & 0.262 & 69 \\
        Prev config + PF across episodes & 0.000 & 0 & {--} & {--} \\
        HWBP + CRF & 0.552 & 83 & 0.773 & 97 \\
        HWBP + PF & 0 & 0 & 0.572 & 87 \\
        \bottomrule
    \end{tabular}
\end{table*}

\subsection{Configuration 3: 9x9 Grid with Internal Partition Wall}

This configuration introduced internal obstacles (a partition wall with a gap for the agent to pass through). For the LLM agent to solve the problem, it needs to break the task into two steps. The first step requires the agent to pass through the partition wall. The second step then requires the LLM agent to reach the target location. Performance metrics are detailed in Table~\ref{tab:merged-config3-results}. In this more complex setting, all models performed quite poorly, though still with noticeable performance decline as additional feedback was provided.

\begin{table*}[t]
    \centering
    \caption{Performance Metrics for Configuration~3.}
    \label{tab:merged-config3-results}
    \begin{tabular}{lSSSS}
        \toprule
        \textbf{Reference Approach} & \textbf{Avg. Cum. Reward} & \textbf{Success Rate (\%)} & & \\ 
        \midrule
        Random Policy & 0.015 & 6 &  &  \\
        RL Baseline & 0.95 & 100 &  &  \\ 
        \midrule
        \textbf{LLM-based Approach} &
        \multicolumn{2}{c}{\textbf{llama3.1:8b}} &
        \multicolumn{2}{c}{\textbf{qwen2.5:1.5b}} \\
        \cmidrule(lr){2-3} \cmidrule(lr){4-5}
        & {\textbf{Avg. Cum. Reward}} & {\textbf{Success Rate (\%)}} 
        & {\textbf{Avg. Cum. Reward}} & {\textbf{Success Rate (\%)}} \\
        \midrule
        HWBP & 0.177 & 31 & 0.17 & 29 \\
        HWBP + CoT & 0.166 & 36 & 0.189 & 35 \\
        HWBP + CoT + DF & 0.129 & 22 & 0 & 0 \\
        HWBP + CoT + DF + RF & 0.104 & 18 & 0 & 0 \\
        HWBP + CoT + DF + RF + CRF & 0.156 & 30 & 0.038 & 10 \\
        HWBP + CoT + DF + RF + CRF + PF & 0.000 & 0 & 0.019 & 20 \\
        Prev config + PF across episodes & 0.000 & 0 & {--} & {--} \\
        HWBP + CRF & 0.134 & 29 &  0.128 &  20 \\
        HWBP + PF & 0 & 0 & 0.071 & 15 \\
        \bottomrule
    \end{tabular}
\end{table*}

\subsection{Results of One-shot Prompting with Reasoning Models}

In Section~\ref{sec:rm}, two reasoning LLMs were selected to evaluate the one-shot performance of the most basic version of the LLM policy, Human-Written-Based-Prompt. The full model outputs are included in Appendix~\ref{app:rm}. Table~\ref{tab:rm} shows that QwQ 32B nearly completed Configuration 1, successfully completed Configuration 2, but failed to generate a plan for Configuration 3. In contrast, Deepseek R1 14B failed on all three configurations.

\begin{table}[h]
    \centering
    \caption{Performance of reasoning models across different configurations.}
    \label{tab:rm}
    \begin{tabular}{lccc}
        \toprule
         & Config 1 & Config 2 & Config 3 \\ 
        \midrule
        Deepseek R1 14B & Failed & Incomplete & Incorrect \\
        QwQ 32B & Nearly complete & Complete & Failed to generate \\
        \bottomrule
    \end{tabular}
\end{table}

These results may, in part, be attributed to the fact that QwQ 32B has more than twice the number of parameters as Deepseek R1 14B and is also a newer model. Interestingly, despite its higher capacity, QwQ 32B was unable to generate a solution for the more complex Config 3. While reasoning models show some promise on less complex tasks, they tend to fail on more complex instances. Additionally, they often generate a significant number of tokens during the “thinking” phase, making them more costly to run and resulting in higher latency compared to their non-reasoning counterparts.

\subsection{Summary of Findings}

Our experimental results reveal that while LLMs can leverage pre-trained knowledge to achieve some zero-shot success—particularly in simpler configurations and with models like llama3.1:8b and qwen2.5:1.5b—they struggle to generalize in more complex environments, highlighting limitations in planning and reasoning without fine-tuning. Feedback mechanisms such as dynamics, reward, and cumulative reward feedback provided limited benefit, especially in complex settings where the LLM had difficulty integrating additional context within the MDP structure. Policy feedback further degraded performance, likely due to the introduction of misleading or irrelevant information. Overall, our findings underscore the limitations of prompt-based methods: simply appending more feedback can dilute the model’s attention, misallocating focus away from critical task-relevant signals and ultimately reducing decision-making effectiveness.

\section{Conclusion}

This work explored the application of large language models (LLMs) for decision-making in Markov Decision Processes (MDPs). By employing structured prompting strategies, the performance of LLMs was evaluated across various configurations of the MiniGrid environment and compared against a proximal policy persuit reinforcement learning (RL) approach.

Our findings reveal that LLM-based strategies, while competitive in simpler scenarios, face significant challenges in complex environments. The inherent advantages of LLMs, such as leveraging pre-trained knowledge, are offset by their limited adaptability and scalability in more demanding settings. Furthermore, while additional feedback mechanisms were intended to enhance decision-making, they often led to performance degradation, especially in environments requiring nuanced reasoning. This is likely to be a challenge in agentic workflows, as they become more complex and tasked with operating on greater levels of information.

\clearpage

\onecolumn
\appendices

\section{Prompt details}\label{app:src}

\begin{lstlisting}[breaklines=true, breakatwhitespace=true, basicstyle=\small\ttfamily,language=Python, caption={prompt\_composer.py}]
import yaml

class PromptComposer:
    def __init__(self, config_file):
        # Read environment-specific variables from the YAML config file
        with open(config_file, 'r') as f:
            config = yaml.safe_load(f)
        
        env_config = config['Environment']
        self.env_name = env_config['env_name']
        self.env_description = env_config['env_description']
        
        # Initialize the constant prompt components
        self.prompt_header = """\
{const_env_name} Environment Instructions:

Your role is to act as a policy for a Markov Decision Process (MDP). You will be provided with a state and asked to produce an action to execute. Your goal is to maximize rewards in the {const_env_name} environment.
"""
        
        self.base_instructions = """\
Instructions:
1. Analyze the Current Situation.
2. Decide on the Best Action to progress toward the goal.
"""

        self.base_instructions_complete = """\
Instructions:
1. Analyze the Current Situation.
2. Decide on the full sequence of actions to reach the goal.
"""
        
        self.cot_instructions = """\
Chain-of-Thought (CoT) Instructions:
1. Analyze the Current Situation:
  - Determine the agent's state
  - Identify any immediate obstacles or challenges facing the agent.
2. Decide on the Best Action:
  - Consider how each action affects the agent and the environment.
  - Choose the action that will optimally satisfy the task.
  - Consider whether it may be worth incurring risk to obtain a higher long term reward.
"""
        
        self.dynamics_feedback = """\
Dynamics Feedback:
The last action you took was {var_prev_action}, which changed the environment state from
{var_pre_state} 
to 
{var_cur_state}.
"""
        
        self.reward_feedback = """\
Reward Feedback:
The agent received a reward of {var_reward} for the last action.
"""
        
        self.cummulative_reward_feedback = """\
Cumulative Reward Feedback:
So far, the agent has taken {var_n} steps and accumulated a total of {var_cummulative_reward} rewards. 
"""

        self.policy_feedback = """\
Policy Feedback:
The previous policy is summarized as: 
{var_previous_policy_summary}
"""
        
        self.output_requirements = """\
Output Requirement: Must provide the selected action in ONLY JSON format (NOT LaTeX) with the following fields:
- "action": The chosen action as a string.
- "reason": A brief explanation of why this action is the best choice.
"""

        self.output_requirements_with_policy_feedback = """\
Output Requirement: Must provide the selected action in ONLY JSON format (NOT LaTeX) with the following fields:
- "action": The chosen action as a string (e.g., "Move Forward").
- "reason": A brief explanation of why this action is the best choice.
- "policy": Summarizes its current policy behind the action selection.
"""

    def compose_prompt(self, include_cot=False, include_dynamics_feedback=False,
                       include_reward_feedback=False, include_cummulative_reward_feedback=False, include_policy_feedback=False, full_instruction=False):
        # Compose the prompt header
        prompt = self.prompt_header.format(const_env_name=self.env_name)
        
        # Add the environment description
        prompt += '\n' + self.env_description
        
        # Add dynamics feedback if included
        if include_dynamics_feedback:
            prompt += '\n' + self.dynamics_feedback
        
        # Add reward feedback if included
        if include_reward_feedback:
            prompt += '\n' + self.reward_feedback
        
        # Add cummulative reward feedback if included
        if include_cummulative_reward_feedback:
            prompt += '\n' + self.cummulative_reward_feedback

        # Add policy feedback if included
        if include_policy_feedback:
            prompt += '\n' + self.policy_feedback
        
        # Add instructions
        if include_cot:
            prompt += '\n' + self.cot_instructions
        else:
            if full_instruction:
                prompt += '\n' + self.base_instructions_complete
            else:
                prompt += '\n' + self.base_instructions
        
        # Add output requirements
        if include_policy_feedback:
            prompt += '\n' + self.output_requirements_with_policy_feedback
        else:
            if not full_instruction:
                prompt += '\n' + self.output_requirements
        
        return prompt

    def get_prompts(self):
        # Generate all prompt versions
        prompts = {}

        prompts['HWBP'] = self.compose_prompt()

        prompts['HWBP_COMPLETE'] = self.compose_prompt(
            full_instruction=True
        )

        prompts['HWBP_PF'] = self.compose_prompt(
            include_policy_feedback=True
        )

        prompts['HWBP_CRF'] = self.compose_prompt(
            include_cummulative_reward_feedback=True
        )

        prompts['HWBP_CoT'] = self.compose_prompt(
            include_cot=True
        )

        prompts['HWBP_CoT_PF'] = self.compose_prompt(
            include_cot=True,
            include_policy_feedback=True
        )

        prompts['HWBP_CoT_DF'] = self.compose_prompt(
            include_cot=True, 
            include_dynamics_feedback=True
        )

        prompts['HWBP_CoT_DF_RF'] = self.compose_prompt(
            include_cot=True, 
            include_dynamics_feedback=True,
            include_reward_feedback=True
        )

        prompts['HWBP_CoT_DF_RF_CRF'] = self.compose_prompt(
            include_cot=True, 
            include_dynamics_feedback=True,
            include_reward_feedback=True, 
            include_cummulative_reward_feedback=True
        )

        prompts['HWBP_CoT_DF_RF_CRF_PF'] = self.compose_prompt(
            include_cot=True, 
            include_dynamics_feedback=True,
            include_reward_feedback=True, 
            include_cummulative_reward_feedback=True,
            include_policy_feedback=True
        )

        prompts['HWBP_CoT_RF_CRF'] = self.compose_prompt(
            include_cot=True, 
            include_reward_feedback=True, 
            include_cummulative_reward_feedback=True
        )
        
        return prompts
\end{lstlisting}.

\begin{lstlisting}[breaklines=true, breakatwhitespace=true, basicstyle=\small\ttfamily, caption={config.yaml}]
Environment:
  env_name: Minigrid
  env_description: |
    In the Minigrid environment, you control an agent navigating through a grid-based world. The objective is to reach the target 'G' while avoiding obstacles and hazards. The grid consists of cells represented by single-character symbols.

    The observation space is the agent's position and facing direction, represented within a grid of symbols in the Minigrid domain. The action space includes movements and interactions that can be performed by the agent to achieve the goal efficiently.

    Agent's Orientation and Movement:
    - The agent can face one of four directions:
        - '>': Facing right
        - 'v': Facing down
        - '<': Facing left
        - '^': Facing up
    - Rotation Actions (changes facing direction, does not move the agent):
        - Turn Left: Rotates the agent 90 degrees to the left (counter-clockwise). Example: From facing up ('^') to facing left ('<').
        - Turn Right: Rotates the agent 90 degrees to the right (clockwise). Example: From facing up ('^') to facing right ('>').
    - Movement Action (changes position in the grid):
        - Move Forward: Moves the agent one cell forward in the direction it is currently facing, provided there is no obstacle.

    Completion Action:
    - Done: Used when the agent has reached the goal and wants to end the game.

    Grid Symbols:
    - 'W': Wall (blocks movement)
    - 'G': Goal
    - ' ': Empty Cell

    The current game state is:
    {var_cur_state}

    Available Actions (Select one action):
    - 'Turn Left'
    - 'Turn Right'
    - 'Move Forward'
    - 'Done'
    
\end{lstlisting}

\begin{lstlisting}[breaklines=true, breakatwhitespace=true, basicstyle=\small\ttfamily, caption={An instance of HWBP + CoT + DF + RF + CRF + PF}]
Minigrid Environment Instructions:

Your role is to act as a policy for a Markov Decision Process (MDP). You will be provided with a state and asked to produce an action to execute. Your goal is to maximize rewards in the Minigrid environment.

In the Minigrid environment, you control an agent navigating through a grid-based world. The objective is to reach the target 'G' while avoiding obstacles and hazards. The grid consists of cells represented by single-character symbols.

The observation space is the agent's position and facing direction, represented within a grid of symbols in the Minigrid domain. The action space includes movements and interactions that can be performed by the agent to achieve the goal efficiently.

Agent's Orientation and Movement:
- The agent can face one of four directions:
    - '>': Facing right
    - 'v': Facing down
    - '<': Facing left
    - '^': Facing up
- Rotation Actions (changes facing direction, does not move the agent):
    - Turn Left: Rotates the agent 90 degrees to the left (counter-clockwise). Example: From facing up ('^') to facing left ('<').
    - Turn Right: Rotates the agent 90 degrees to the right (clockwise). Example: From facing up ('^') to facing right ('>').
- Movement Action (changes position in the grid):
    - Move Forward: Moves the agent one cell forward in the direction it is currently facing, provided there is no obstacle.

Completion Action:
- Done: Used when the agent has reached the goal and wants to end the game.

Grid Symbols:
- 'W': Wall (blocks movement)
- 'G': Goal
- ' ': Empty Cell

The current game state is:
{var_cur_state}

Available Actions (Select one action):
- 'Turn Left'
- 'Turn Right'
- 'Move Forward'
- 'Done'

Dynamics Feedback:
The last action you took was {var_prev_action}, which changed the environment state from
{var_pre_state} 
to 
{var_cur_state}.

Reward Feedback:
The agent received a reward of {var_reward} for the last action.

Cumulative Reward Feedback:
So far, the agent has taken {var_n} steps and accumulated a total of {var_cummulative_reward} rewards. 

Policy Feedback:
The previous policy is summarized as: 
{var_previous_policy_summary}

Chain-of-Thought (CoT) Instructions:
1. Analyze the Current Situation:
  - Determine the agent's state
  - Identify any immediate obstacles or challenges facing the agent.
2. Decide on the Best Action:
  - Consider how each action affects the agent and the environment.
  - Choose the action that will optimally satisfy the task.
  - Consider whether it may be worth incurring risk to obtain a higher long term reward.

Output Requirement: Must provide the selected action in ONLY JSON format (NOT LaTeX) with the following fields:
- "action": The chosen action as a string (e.g., "Move Forward").
- "reason": A brief explanation of why this action is the best choice.
\end{lstlisting}

\newpage

\section{Hyperparameters for Reinforcement Learning}\label{app:rl_hyper}

\begin{table*}[h!]
\centering
\begin{tabular}{ll}
\textbf{Component} & \textbf{Value / Setting} \\
\hline
RL Algorithm & PPO (\texttt{stable\_baselines3.PPO}) \\
Policy name & \texttt{"CnnPolicy"} \\
Feature Extractor & \texttt{MinigridFeaturesExtractor} (custom CNN + FC layer) \\
CNN Architecture & Conv2D layers with 16 → 32 → 64 filters, kernel size (2,2), ReLU, Flatten \\
Feature Dim & 128 via \texttt{features\_extractor\_kwargs} \\
Learning Rate & 2.5e-4 \\
\end{tabular}
\caption{Shared settings across all configurations.}\label{tab:rl_shared_params}
\end{table*}

\begin{table*}[h!]
\centering
\begin{tabular}{llll}
\textbf{Parameter} & \textbf{Config 1} & \textbf{Config 2} & \textbf{Config 3} \\
\hline
Environment & \texttt{MiniGrid-Empty-Random-5x5-v0} & \texttt{MiniGrid-Empty-16x16-v0} & \texttt{MiniGrid-SimpleCrossingS9N1-v0} \\
Environment Size & Small (5x5) & Large (16x16) & Medium (9x9) \\
Training Steps & 30,000 & 300,000 & 3,000,000 \\
\end{tabular}
\caption{Key differences across configurations.}\label{tab:rl_differences}
\end{table*}

\newpage

\section{One-shot prompting with reasoning models}\label{app:rm}

\begin{figure*}[ht]
\centering
\begin{tikzpicture}[node distance=0.6cm]
  \tikzstyle{user} = [rectangle, rounded corners, draw=blue, fill=blue!10, text width=0.9\textwidth, minimum height=1cm, align=left]
  \tikzstyle{ai} = [rectangle, rounded corners, draw=green!50!black, fill=green!10, text width=0.9\textwidth, minimum height=1cm, align=left]

  \node[user] (u1) {User: \\
\textlangle Description of the problem (see figure \ref{fig:minigrid-configs}, section \ref{sec:llm_policies}, and appendix \ref{app:src})\textrangle \\
Instructions: \\
1. Analyze the Current Situation. \\
2. Decide on the full sequence of actions to reach the goal.};
  \node[ai, below=of u1] (a1) {Deepseek R1-14B: \\
\textlangle Thinking process\textrangle \\
The optimal path for the agent starts by turning right to face downwards and then moving directly towards the goal. This approach efficiently navigates the grid with minimal steps. \\
**Step-by-Step Explanation:** \\
1. **Initial Position:** The agent begins at position (1,1) facing upwards. \\
2. **Turn Right:** The agent turns right, now facing downward ('v'). \\
3. **Move Down Twice:** The agent moves down from (1,1) to (2,1), then to (3,1). \\
4. **Turn Right Again:** Now at (3,1), the agent turns right to face right ('$>$'). \\
5. **Move Right Twice:** From (3,1), the agent moves right to (3,2), then to (3,3), which is the goal. \\
6. **Done:** The agent reaches the goal. \\
This sequence ensures the agent takes the shortest path by leveraging right turns and downward movements. \\
**Answer:** \\
The optimal actions for the agent are: RIGHT, DOWN, DOWN, RIGHT, RIGHT. \\
These steps guide the agent from (1,1) to (3,3), achieving the goal efficiently.};
  \node[ai, below=of a1] (a2) {QwQ 32B: \\
\textlangle Thinking process\textrangle \\ 
The problem requires finding the shortest sequence of moves for an agent to reach a target position, considering each move (forward) and turn costs one action. The agent starts facing east at position (1,1). Assuming the target is two units to the right and one unit down from the starting position: \\
1. **Move East twice**: This takes 2 actions (FF). \\
2. **Turn Right** (south): Costs 1 action (R). \\
3. **Move South once**: Costs 1 action (F). \\
This results in the sequence FFRF, which is four actions.};

\end{tikzpicture}

\caption{Configuration 1.}
\end{figure*}

\begin{figure*}[ht]
\centering
\begin{tikzpicture}[node distance=0.6cm]
  \tikzstyle{user} = [rectangle, rounded corners, draw=blue, fill=blue!10, text width=0.9\textwidth, minimum height=1cm, align=left]
  \tikzstyle{ai} = [rectangle, rounded corners, draw=green!50!black, fill=green!10, text width=0.9\textwidth, minimum height=1cm, align=left]

  \node[user] (u1) {User: \\
\textlangle Description of the problem (see figure \ref{fig:minigrid-configs}, section \ref{sec:llm_policies}, and appendix \ref{app:src})\textrangle \\
Instructions: \\
1. Analyze the Current Situation. \\
2. Decide on the full sequence of actions to reach the goal.};
  \node[ai, below=of u1] (a1) {Deepseek R1-14B: \\
\textlangle Thinking process\textrangle \\
The optimal first action for the agent is to 'Move Forward' as there are no obstacles directly in front of it, allowing it to advance towards the goal. \\
**Step-by-Step Explanation:** \\
1. **Current State Analysis:** The agent is at (row=1, column=1), facing right ('$>$'). The cell immediately ahead (row=1, column=2) is empty (' '), so moving forward is possible without collision. \\
2. **Action Selection:** Among the available actions, 'Move Forward' is feasible and beneficial as it directly progresses the agent towards the goal without hindrance. \\
**Answer:** \\
The agent should execute the action: \\
- Move Forward};
  \node[ai, below=of a1] (a2) {QwQ 32B: \\
\textlangle Thinking process\textrangle \\ 
To solve this problem, we need to guide a robot from the top-left corner of a grid to the bottom-right corner using the fewest moves possible. The robot can move forward until it hits a wall and can turn directions. Assuming the robot starts facing east: \\
1. **Initial Setup**: The robot begins at the top-left (0, 0) facing east. \\
2. **First Move**: Move straight east to the end of the row. This will take the robot to the easternmost point in that row without obstacles. \\
3. **Turn Direction**: Once at the eastern edge, turn south. \\
4. **Second Move**: Move straight south until reaching the bottom-right corner (the target), as moving south from the top row's east end directly leads to the destination. \\
This approach leverages the robot's ability to move in a straight line until it hits a wall, minimizing the number of commands needed. \\
```python \\
def navigate\_robot(): \\
\quad \# Turn east and move all the way east \\
\quad turn\_east() \\
\quad move() \\
\quad \# Turn south and move all the way south \\
\quad turn\_south() \\
\quad move() \\
``` \\
**Explanation**: \\
- **Turn East**: Ensures the robot faces right. \\
- **Move East**: The robot moves to the easternmost point in its row. \\
- **Turn South**: Redirects the robot downward. \\
- **Move South**: The robot proceeds straight down to the bottom-right corner. \\
This sequence efficiently navigates the robot using two turns and two moves.};

\end{tikzpicture}
\caption{Configuration 2.}
\end{figure*}

\begin{figure*}[ht]
\centering
\begin{tikzpicture}[node distance=0.6cm]
  \tikzstyle{user} = [rectangle, rounded corners, draw=blue, fill=blue!10, text width=0.9\textwidth, minimum height=1cm, align=left]
  \tikzstyle{ai} = [rectangle, rounded corners, draw=green!50!black, fill=green!10, text width=0.9\textwidth, minimum height=1cm, align=left]

  \node[user] (u1) {User: \\
\textlangle Description of the problem (see figure \ref{fig:minigrid-configs}, section \ref{sec:llm_policies}, and appendix \ref{app:src})\textrangle \\
Instructions: \\
1. Analyze the Current Situation. \\
2. Decide on the full sequence of actions to reach the goal.};
  \node[ai, below=of u1] (a1) {Deepseek R1-14B: \\
\textlangle Thinking process\textrangle \\
To solve this problem, I would typically use a Breadth-First Search (BFS) algorithm to find the shortest path in a grid from the starting position to the goal. However, since the specific details of the grid aren't provided, I'll outline a general approach and provide an example solution based on a standard grid setup. \\
**General Approach:** \\
1. **Problem Analysis:** Understand the grid layout, identify start and goal positions, and note any obstacles or walls. \\
2. **Algorithm Selection:** Use BFS for shortest path in unweighted grids due to its ability to explore all possible moves level by level. \\
3. **Implementation Steps:** \\
   - Initialize a queue with the starting position. \\
   - Mark visited positions to avoid revisiting them. \\
   - For each position, explore all possible directions (up, down, left, right). \\
   - If the goal is reached, reconstruct the path. \\
**Example Solution (Assuming a Simple Grid):** \\
Let's assume a grid where: \\
- Start = (1, 1) \\
- Goal = (7, 8) \\
- Walls are present at specific positions as in a standard Minigrid setup. \\
**Step-by-Step Explanation:** \\
1. **Initialization:** \\
\quad - Queue starts with [(1, 1)]. \\
\quad - Visited set contains (1, 1). \\
2. **BFS Execution:** \\
\quad - Dequeue (1, 1). \\
\quad - Explore neighbors: \\
\quad\quad - Right to (2, 1) if not a wall and not visited. \\
\quad\quad - Down to (1, 2) if valid. \\
\quad - Enqueue these positions and mark them as visited. \\
3. **Continue BFS:** \\
\quad - Process each position, exploring all possible moves while avoiding walls and revisits.
\quad - Eventually reach the goal by moving right across y=1 until x=7, then down to y=8.
**Final Answer (Example Path):** \\
The shortest path from (1, 1) to (7, 8) is: \\
``` \\
(1,1) → (2,1) → (3,1) → (4,1) → (5,1) → (6,1) → (7,1) → (7,2) → ... → (7,8) \\
``` \\
However, the exact path depends on the grid's walls. This example assumes a clear rightward and downward path. \\
For precise solutions, implement BFS with the actual grid data.};
  \node[ai, below=of a1] (a2) {QwQ 32B: \\
\textlangle Thinking process\textrangle};

\end{tikzpicture}
\caption{Configuration 3.}
\end{figure*}


\begin{thebibliography}{10}
\providecommand{\url}[1]{\texttt{#1}}
\providecommand{\urlprefix}{URL }
\providecommand{\doi}[1]{https://doi.org/#1}

\bibitem{ollama_qwen25_2024}
{Alibaba}: Qwen 2.5 (2024), \url{https://ollama.com/library/qwen2.5}, accessed: 2025-05-22

\bibitem{ollama_qwq32b_2025}
{Alibaba}: qwq:32b (2025), \url{https://ollama.com/library/qwq:32b}, accessed: 2025-05-27

\bibitem{chevalier2023minigrid}
Chevalier-Boisvert, M., Dai, B., Towers, M., Perez-Vicente, R., Willems, L., Lahlou, S., Pal, S., Castro, P.S., Terry, J.: Minigrid \& miniworld: Modular \& customizable reinforcement learning environments for goal-oriented tasks. Advances in Neural Information Processing Systems  \textbf{36},  73383--73394 (2023)

\bibitem{darwiche2024llms}
Darwiche, A.: Can llms really reason and plan? (2024), \url{https://cacm.acm.org/blogcacm/can-llms-really-reason-and-plan/}, communications of the ACM Blog, accessed June 22, 2025

\bibitem{ollama_deepseekr114b_2025}
{Deepseek}: deepseek-r1:14b (2025), \url{https://ollama.com/library/deepseek-r1:14b}, accessed: 2025-05-27

\bibitem{grattafiori2024llama}
Grattafiori, A., Dubey, A., Jauhri, A., Pandey, A., Kadian, A., Al-Dahle, A., Letman, A., Mathur, A., Schelten, A., Vaughan, A., et~al.: The llama 3 herd of models. arXiv preprint arXiv:2407.21783  (2024)

\bibitem{guan2023leveraging}
Guan, L., Valmeekam, K., Sreedharan, S., Kambhampati, S.: Leveraging pre-trained large language models to construct and utilize world models for model-based task planning. Advances in Neural Information Processing Systems  \textbf{36},  79081--79094 (2023)

\bibitem{guo2025deepseek}
Guo, D., Yang, D., Zhang, H., Song, J., Zhang, R., Xu, R., Zhu, Q., Ma, S., Wang, P., Bi, X., et~al.: Deepseek-r1: Incentivizing reasoning capability in llms via reinforcement learning. arXiv preprint arXiv:2501.12948  (2025)

\bibitem{huang2022language}
Huang, W., Abbeel, P., Pathak, D., Mordatch, I.: Language models as zero-shot planners: Extracting actionable knowledge for embodied agents. In: International conference on machine learning. pp. 9118--9147. PMLR (2022)

\bibitem{huang2025ai}
Huang, Z.: The ai revolution in distributed systems: How github copilot verified azure storage (2025), \url{\url{https://zfhuang99.github.io/github%20copilot/formal%20verification/tla+/2025/05/24/ai-revolution-in-distributed-systems.html}}, accessed: 2025-06-26

\bibitem{ju2024globe}
Ju, D., Jiang, S., Cohen, A., Foss, A., Mitts, S., Zharmagambetov, A., Amos, B., Li, X., Kao, J.T., Fazel-Zarandi, M., et~al.: To the globe (ttg): Towards language-driven guaranteed travel planning. arXiv preprint arXiv:2410.16456  (2024)

\bibitem{kambhampati2025stop}
Kambhampati, S., Stechly, K., Valmeekam, K., Saldyt, L., Bhambri, S., Palod, V., Gundawar, A., Samineni, S.R., Kalwar, D., Biswas, U.: Stop anthropomorphizing intermediate tokens as reasoning/thinking traces! arXiv preprint arXiv:2504.09762  (2025)

\bibitem{kambhampati2024position}
Kambhampati, S., Valmeekam, K., Guan, L., Verma, M., Stechly, K., Bhambri, S., Saldyt, L.P., Murthy, A.B.: Position: Llms can’t plan, but can help planning in llm-modulo frameworks. In: Forty-first International Conference on Machine Learning (2024)

\bibitem{khattab2023dspy}
Khattab, O., Singhvi, A., Maheshwari, P., Zhang, Z., Santhanam, K., Vardhamanan, S., Haq, S., Sharma, A., Joshi, T.T., Moazam, H., et~al.: Dspy: Compiling declarative language model calls into self-improving pipelines. arXiv preprint arXiv:2310.03714  (2023)

\bibitem{ollama_llama3_2024}
{Meta Platforms}: Llama 3.1 (2024), \url{https://ollama.com/library/llama3.1}, accessed: 2024-10-10

\bibitem{prakash2023llm}
Prakash, B., Oates, T., Mohsenin, T.: Llm augmented hierarchical agents. arXiv preprint arXiv:2311.05596  (2023)

\bibitem{raschka2025understanding}
Raschka, S.: Understanding reasoning in large language models. \url{https://sebastianraschka.com/blog/2025/understanding-reasoning-llms.html} (2025), accessed: 2025-06-18

\bibitem{schulman2017proximal}
Schulman, J., Wolski, F., Dhariwal, P., Radford, A., Klimov, O.: Proximal policy optimization algorithms. arXiv preprint arXiv:1707.06347  (2017)

\bibitem{schultz2024mastering}
Schultz, J., Adamek, J., Jusup, M., Lanctot, M., Kaisers, M., Perrin, S., Hennes, D., Shar, J., Lewis, C., Ruoss, A., et~al.: Mastering board games by external and internal planning with language models. arXiv preprint arXiv:2412.12119  (2024)

\bibitem{shojaee2025illusion}
Shojaee, P., Mirzadeh, I., Alizadeh, K., Horton, M., Bengio, S., Farajtabar, M.: The illusion of thinking: Understanding the strengths and limitations of reasoning models via the lens of problem complexity. arXiv preprint arXiv:2506.06941  (2025)

\bibitem{stechly2024self}
Stechly, K., Valmeekam, K., Kambhampati, S.: On the self-verification limitations of large language models on reasoning and planning tasks. arXiv preprint arXiv:2402.08115  (2024)

\bibitem{tang2024mba}
Tang, X., Gao, Q., Li, J., Du, N., Li, Q., Xie, S.: Mba-rag: a bandit approach for adaptive retrieval-augmented generation through question complexity. arXiv preprint arXiv:2412.01572  (2024)

\bibitem{valmeekam2023planbench}
Valmeekam, K., Marquez, M., Olmo, A., Sreedharan, S., Kambhampati, S.: Planbench: An extensible benchmark for evaluating large language models on planning and reasoning about change. Advances in Neural Information Processing Systems  \textbf{36},  38975--38987 (2023)

\bibitem{valmeekam2023planning}
Valmeekam, K., Marquez, M., Sreedharan, S., Kambhampati, S.: On the planning abilities of large language models-a critical investigation. Advances in Neural Information Processing Systems  \textbf{36},  75993--76005 (2023)

\bibitem{wei2022chain}
Wei, J., Wang, X., Schuurmans, D., Bosma, M., Xia, F., Chi, E., Le, Q.V., Zhou, D., et~al.: Chain-of-thought prompting elicits reasoning in large language models. Advances in neural information processing systems  \textbf{35},  24824--24837 (2022)

\bibitem{yang2025qwen3}
Yang, A., Li, A., Yang, B., Zhang, B., Hui, B., Zheng, B., Yu, B., Gao, C., Huang, C., Lv, C., et~al.: Qwen3 technical report. arXiv preprint arXiv:2505.09388  (2025)

\bibitem{yao2023react}
Yao, S., Zhao, J., Yu, D., Du, N., Shafran, I., Narasimhan, K., Cao, Y.: React: Synergizing reasoning and acting in language models. In: International Conference on Learning Representations (ICLR) (2023)

\bibitem{yuksekgonul2024textgrad}
Yuksekgonul, M., Bianchi, F., Boen, J., Liu, S., Huang, Z., Guestrin, C., Zou, J.: Textgrad: Automatic" differentiation" via text. arXiv preprint arXiv:2406.07496  (2024)

\end{thebibliography}
\end{document}